\documentclass[10pt,twocolumn,letterpaper]{article}

\usepackage{cvpr}              

\usepackage{url}            
\usepackage{xurl}
\usepackage{booktabs}       
\usepackage{amsfonts}       
\usepackage{nicefrac}       

\usepackage{xcolor}         
\usepackage{ulem}
\usepackage{pifont}
\usepackage{utfsym}

\usepackage{siunitx}
\usepackage{adjustbox}

\usepackage{makecell}
\usepackage{multirow}
\usepackage{graphicx} 
\usepackage{tcolorbox}
\usepackage{listings}

\usepackage{microtype}      

\definecolor{cvprblue}{rgb}{0.21,0.49,0.74}
\usepackage[pagebackref,breaklinks,colorlinks,allcolors=cvprblue]{hyperref}

\def\paperID{9964} 
\def\confName{CVPR}
\def\confYear{2026}

\title{Physical Simulator In-the-Loop Video Generation}

\author{
    Lin Geng Foo$^{1,5}$ \quad
    Mark He Huang$^{2,3}$ \quad
    Alexandros Lattas$^{4}$ \quad
    Stylianos Moschoglou$^{4}$ \\
    Thabo Beeler$^{4}$ \quad
    Christian Theobalt$^{1,5}$ \\[2mm]
    {\small $^{1}$Max Planck Institute for Informatics, Saarland Informatics Campus} \quad
    {\small$^{2}$Singapore University of Technology and Design} \\
    {\small$^{3}$A*STAR} \quad    
    {\small$^{4}$Google} \quad
    {\small$^{5}$Saarbrücken Research Center for Visual Computing, Interaction and Artificial Intelligence}
}

\begin{document}
\maketitle

\begin{abstract}
Recent advances in diffusion-based video generation have achieved remarkable visual realism but still struggle to obey basic physical laws such as gravity, inertia, and collision. Generated objects often move inconsistently across frames, exhibit implausible dynamics, or violate physical constraints, limiting the realism and reliability of AI-generated videos.
We address this gap by introducing Physical Simulator In-the-loop Video Generation (PSIVG), a novel framework that integrates a physical simulator into the video diffusion process. Starting from a template video generated by a pre-trained diffusion model, PSIVG reconstructs the 4D scene and foreground object meshes, initializes them within a physical simulator, and generates physically consistent trajectories. These simulated trajectories are then used to guide the video generator toward spatio-temporally physically coherent motion. 
To further improve texture consistency during object movement, we propose a Test-Time Texture Consistency Optimization (TTCO) technique that adapts text and feature embeddings based on pixel correspondences from the simulator.
Comprehensive experiments demonstrate that PSIVG produces videos that better adhere to real-world physics while preserving visual quality and diversity.
Project Page: 
\href{https://vcai.mpi-inf.mpg.de/projects/PSIVG/}{https://vcai.mpi-inf.mpg.de/projects/PSIVG}
\end{abstract}

\section{Introduction}

The generation of physically consistent videos represents a key frontier at the intersection of computer vision, graphics, and physical simulation.
If achieved, physical consistency significantly enhances visual realism in generated videos, making AI-generated content more compelling for various commercial applications such as film production, virtual reality, and gaming. 
Ensuring adherence to physics also improves reliability in safety-critical domains like robotics and autonomous driving \cite{wang2025generative}, especially in the use of AI-generated videos to train agent models, directly contributing to the agents’ ability to make sound decisions in real-world settings. 
Due to its widespread importance, this research direction has recently attracted much attention \cite{li2025pisa,liu2024physgen,motamed2025generative}.

\begin{figure}[t]
    \centering
    \includegraphics[width=0.48\textwidth]{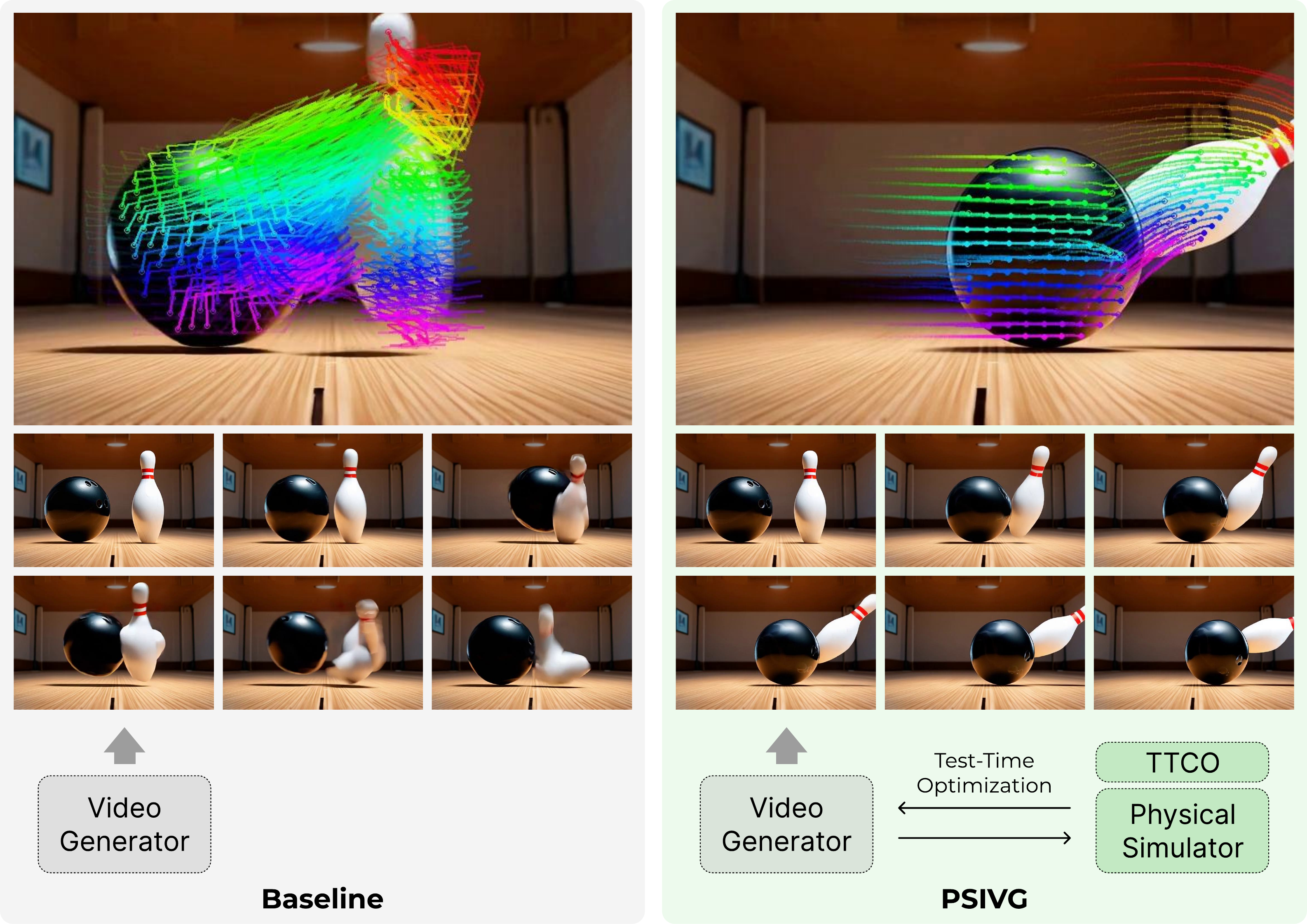}
    \vspace{-4.5mm}
    \caption{
    A baseline video generator (left) produces a physically implausible bowling collision with chaotic motion vectors across the video frames. Our PSIVG framework (right) integrates a physical simulator into the generation loop, guiding the video generator to produce a physically plausible and temporally coherent video.}
    \vspace{-4mm}
    \label{fig:physicalplausibilityissues}
\end{figure}

Over recent years, the quality of generated videos \cite{yang2025cogvideox,kong2024hunyuanvideo,foo2025ai} has improved tremendously, largely driven by diffusion models and large-scale training. However, even the most advanced video generation models still struggle to capture fundamental physics. Generated scenes often contain objects that lack 3D consistency throughout the frames, vanish abruptly, or move in ways that violate real-world physical laws 
(see Fig.~\ref{fig:physicalplausibilityissues} for an example).
Furthermore, crucial physical principles such as gravity, inertia, and collisions are frequently ignored or inaccurately represented. These shortcomings in physical realism have been increasingly observed in recent studies \cite{motamed2025generative, kang2025how}, highlighting a critical gap between visual fidelity and physical plausibility.

We observe that a main reason for this is that modern video generation models are often trained on denoising or reconstruction objectives, which mainly encourages the models to ``denoise'' individual pixels or patches, and thus lack an explicit understanding of physics since there is no mechanism to enforce physical constraints. 
To address this, we propose to integrate physical simulators into video diffusion models -- a paradigm we refer to as simulation-in-the-loop generation. 
The simulator serves as a physics-aware constraint, guiding the diffusion model to maintain consistency across time and space. 
We explore the following research question:
How can we effectively incorporate information from the physical simulator into the video diffusion process to achieve physically consistent generation?

In this work, we  propose Physical Simulator In-the-loop Video Generation (PSIVG), a novel method for physically consistent video generation from text prompts. 
Firstly, we generate a \textit{template video} using a pre-trained video generator, which generates the scene background, the camera movements within the scene, the objects, as well as the initial movements which we will use. 
However, the template video is not physically consistent.
To enforce physical consistency, our method relies on a \textit{physical simulator in-the-loop}, which produces physically consistent trajectories for the objects in the video. 
To incorporate a physical simulator, we first design a perception pipeline (Sec. \ref{sec:perceptionpipeline}) that approximately predicts the 3D meshes of foreground objects as well as reconstructs the 4D scene from the template video.
Next, this information is used to initialize the scene in the physical simulator (Sec. \ref{sec:physicalsimulation}), including placing and scaling the objects, inferring their physical properties, as well as the initial velocity and rotation of each object.
Then, by extracting the RGB, segmentation masks, and pixel-correspondences from the physical simulator, we use them to guide the video generation model to generate physically consistent outputs (Sec. \ref{sec:videogen}).

However, we found that directly using the outputs of the physical simulator as conditional inputs for the video generation model \cite{burgert2025go,yang2025cogvideox} is often not enough for high-quality video generation.
Specifically, we find that the texture of the objects are often not consistent across the frames, where there is flickering or discoloration of objects during movements and rotations.
Texture flickering not only reduces visual quality, but also breaks the perceived temporal coherence necessary for physical realism.
To address this, we further design a \textit{test-time texture-consistency optimization} (TTCO) technique to better maintain the textures of moving objects.
Intuitively, we optimize learnable parameters such that the generated video more closely follows the pixel-to-pixel correspondences from the physical simulator, improving the texture-consistency during movements and rotations.
To achieve a localized optimization targeting the moving foreground objects while maintaining the background, we optimize text embeddings and features corresponding to the foreground objects.
TTCO enables the stronger incorporation of physics constraints into the video diffusion model, even when using pre-trained, open-source models that may not be effective at enforcing physical consistency.
We highlight that no additional training data is required for our TTCO.

In summary, our contributions are:
1) We propose PSIVG, a novel physical simulator in-the-loop video generation pipeline. PSIVG is the first training-free, inference-time framework to bridge a generative text-to-video pipeline with a 3D physical simulator, enabling on-the-fly physical-consistency guidance for pre-trained video diffusion models.
2) To incorporate the physical simulator in the loop, we design a perception pipeline that reconstructs 3D object meshes and 4D scene motion for initializing the simulator.
3) To further improve the texture consistency of the moving foreground objects, we introduce TTCO, a test-time optimization strategy that improves texture consistency of moving objects guided by simulator correspondences.

\section{Related Work}

\textbf{Video Generation Models.}
Video generation has recently attracted significant attention both in academia and industry \cite{foo2025ai,Veo3,Sora2,Metamoviegen}.
Most recent approaches are based on video diffusion models \cite{ho2022video,ho2020denoising}, enabling text-to-video \cite{kong2024hunyuanvideo,yang2025cogvideox} and image-to-video generation \cite{blattmann2023stable,yang2025cogvideox}.
To improve controllability, methods incorporate additional inputs such as masks \cite{akkerman2025interdyn,zhang2025vhoi} or multi-modal spatial cues such as masks, depth, and edges \cite{alhaija2025cosmos}.
Other works guide motion generation using trajectories \cite{geng2025motion,ling2025motionclone,namekata2025sgiv} or optical flow \cite{burgert2025go}.
Despite rapid progress, achieving physical consistency in generated videos remains difficult, likely due to the inadequacy of reconstruction losses in learning physical principles.
We build upon recent diffusion-based generators \cite{burgert2025go,yang2025cogvideox} and introduce physically consistent control signals from a physics simulator.
To the best of our knowledge, this is the first approach that integrates a physical simulator in-the-loop into a text-to-video diffusion-based generation pipeline.
We further enhance quality through a test-time texture-consistency optimization.

\begin{figure*}[t]
    \centering
  \includegraphics[width=1\linewidth]{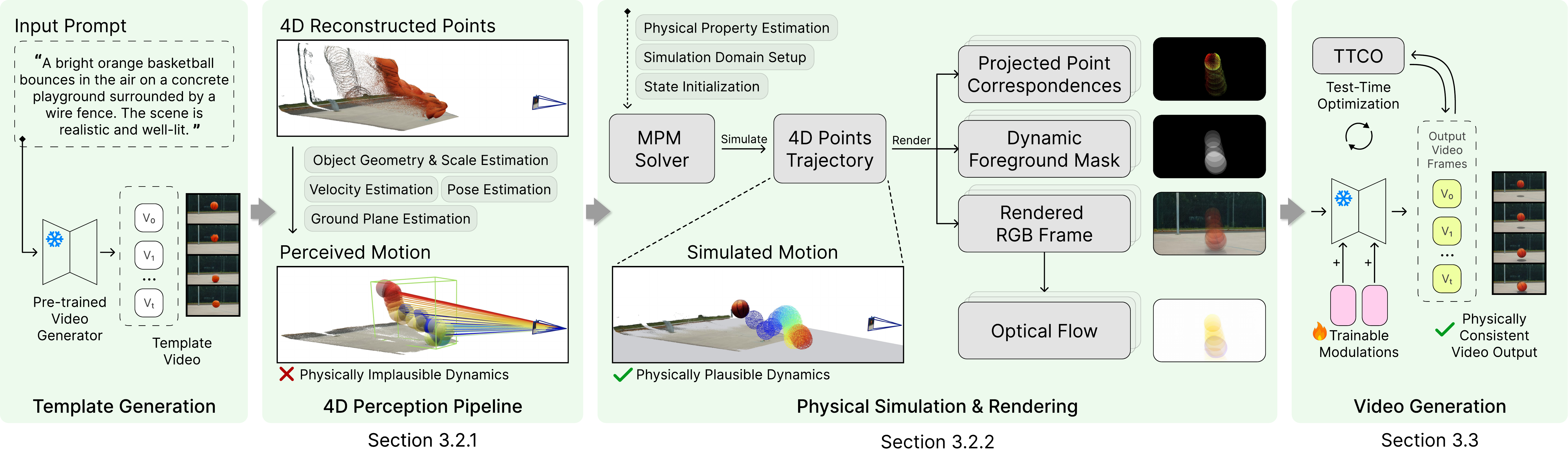}
\caption{
Overview of our Physical Simulator In-the-loop Video Generation (PSIVG) framework.
From an input prompt, a template video is first generated, and is processed by our perception pipeline (Sec. \ref{sec:perceptionpipeline}). The outputs of the perception pipeline are further processed before being passed into the physical simulator (Sec. \ref{sec:physicalsimulation}). 
The rendered outputs from the simulator are then used for video generation (Sec. \ref{sec:videogen}), and this video generation can be improved with TTCO (see Sec. \ref{sec:testtimeopt}) for better texture consistency.
}
\label{fig:pipeline}
\end{figure*}

\textbf{Physically Consistent Generation.}
Several works have explored physics-aware generative models.
Early efforts couple image-based simulators with image generators \cite{montanaro2024motioncraft,liu2024physgen}, but these rely on simplified 2D rigid-body assumptions (e.g., spheres, cones), limiting 3D understanding and temporal texture coherence.
More recently, a line of works \cite{xie2025physanimator,chen2025physgen3d,li2025wonderplay} takes input images and user actions (such as manual 2D strokes or rigging) to generate videos.
PhysAnimator \cite{xie2025physanimator} focuses  on animating cartoons, extracting 2D meshes and applying a 2D simulator, then rendering with a fine-tuned sketch-guided diffusion model.
PhysGen3D \cite{chen2025physgen3d} focuses on obtaining a 3D representation for MPM simulation from input images, where forces can then be applied to this simulatable representation to render video sequences from the physical simulator.
Differently, we handle open-vocabulary video generation, necessitating our 4D perception pipeline to recover physical states (e.g., object rotations), camera motion, and geometry from a generated template video. Our approach of incorporating a pre-trained video diffusion model also provides better video quality and robustness, effectively tolerating reconstruction errors, e.g., errors at the back of reconstructed objects are now refined by the video diffusion model and our TTCO.
WonderPlay \cite{li2025wonderplay} first generates a 3D Gaussian surfel scene from a single image, then use generated videos to supervise updating of the 3D scene, such that videos can be rendered from the 3D scene.
Meanwhile, we perform video refinement directly with our TTCO, which is simpler and more efficient than optimizing a 3D scene; we also support stable 3D object rotations via reconstructed 3D object geometries, which are hard to achieve for a fitted Gaussian surfel scene.
For text-to-video generation, some recent studies impose physical consistency via text-driven or LLM-based reasoning: \cite{xue2025phyt2v} employs physics-grounded prompts, while \cite{lv2024gpt4motion} generates Blender scripts to guide scene construction; such LLM-based explorations are orthogonal to our work.
Besides, PISA \cite{li2025pisa} learns from simulated object interactions and introduces the PisaBench benchmark.
Concurrent works also fine-tune diffusion models using physical forces \cite{gillman2025force} or simulator-derived parameters \cite{wang2025physctrl}.
In contrast, our approach is training-free and does not require additional data, and instead embeds a physical simulator directly within the text-to-video generation loop, enforcing physically consistent dynamics while preserving the high visual fidelity of diffusion-based synthesis.

\textbf{Physical Simulators.}
Physical simulators provide controlled environments for modeling object interactions under physically accurate dynamics.
Widely used engines such as PyBullet \cite{coumans2019} and MuJoCo \cite{todorov2012mujoco} support rigid-body dynamics, collision detection, and robotic control, making them popular in reinforcement learning research.
Beyond rigid-body simulation, methods based on the Material Point Method (MPM) \cite{stomakhin2013material} allow for realistic modeling of deformable materials and have been implemented in frameworks such as Taichi \cite{hu2019taichi} and Warp~\cite{warp2022,10.1145/3610548.3618207}.
Recent efforts also bridge simulation and rendering, for example via 3D Gaussian-based representations \cite{xie2024physgaussian}, and accelerate computation with GPU-optimized engines such as Genesis \cite{Genesis}.
Yet, while these simulators are physically accurate, they lack generative capabilities and depend on predefined 3D assets and material properties.
Moreover, they often cannot capture complex details such as fine-grained textures, lighting, or fluid dynamics.
In contrast, our method couples a physical simulator with a video generative  model, combining the physical accuracy of simulation with the visual realism of diffusion models.
This enables video generation that is both physically grounded and visually compelling.

\section{Method}

\subsection{PSIVG Pipeline Overview}
\label{sec:fullpipeline}

Our goal is to generate videos whose object motions respect real-world physics while maintaining high visual fidelity. To enable this, we introduce a Physical Simulator In-the-loop Video Generation (PSIVG) framework that integrates physics simulation guidance into a pre-trained video diffusion model.
Given an input text prompt, we first generate a \textit{template video} using a pre-trained video generator. 
This sampled \textit{template video},
although typically problematic in following physical laws,
provides essential scene attributes, such as scene composition, camera movements, objects' geometry and textures. 
Next, we invoke our \textit{perception pipeline} to lift these scene attributes from 2D to 3D and acquire information about its intended 4D dynamics.
Using these scene attributes, 
we initialize a physical simulator and perform forward simulation to obtain physically plausible object trajectories.
Finally, the simulator outputs are rendered and fed back into a \textit{video generator}, guiding it to produce videos that follow a physically consistent motion. 
To further improve texture consistency, we propose a test-time texture-consistency optimization (TTCO) technique (Sec.~\ref{sec:videogen}).
TTCO enables physics-aware refinement without requiring additional data for retraining the model.
Refer to Fig.~\ref{fig:pipeline} for an overview.

\subsection{Incorporating the Simulator In-the-loop}
\label{sec:simulatorintheloop}

To incorporate the physical simulator into the generation loop, we do the following after generating the template video:
First, we run our \textit{perception pipeline} to obtain 4D scene elements and dynamics from the template video, namely, 3D foreground/background geometries, active object motions, and camera trajectories, which are required for the physical simulation.
Next, we run the \textit{physical simulation}, which involves setting up the scene, placing the objects, inferring physical properties, initializing the starting state (e.g., velocity, rotations) and the camera movements.
Then, we render the simulated scene and compute physically accurate motion cues as guidance for the video generator, which obey physical principles such as gravity, inertia, and collisions.

\subsubsection{Perception Pipeline}
\label{sec:perceptionpipeline}

\begin{figure}[t] 
  \centering
  \includegraphics[width=\linewidth]{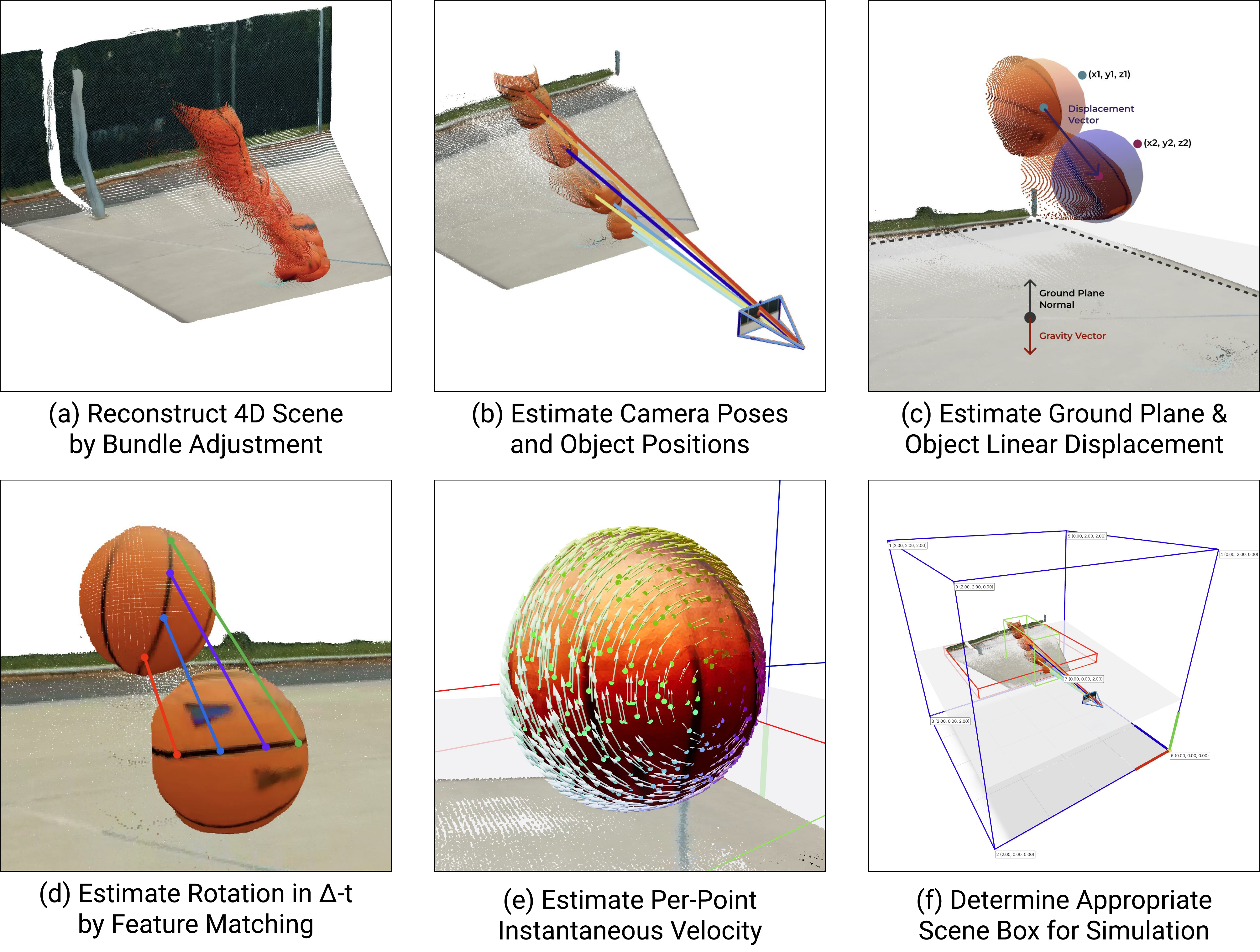}
  \vspace{-4mm}
  \caption{Visualization of the sub-steps in our perception.}  

  \label{fig:perceptionfigs}
  \vspace{-3mm}
\end{figure}

The primary objective of the perception pipeline is to transform a generated template video into simulator-ready assets, thereby bridging the gap between the physical simulator and the generative capabilities of video models.
To enable physically accurate guidance within the simulator, it is essential to extract three key components from the scene: (1) The dynamics of foreground moving objects; (2) the physical environment with which these objects interact; (3) the camera motion.
However, accurately decomposing these components from a generated template video is highly challenging and inherently ill-posed. This difficulty arises because videos produced by vanilla generative models struggle to maintain object permanence and geometric consistency across both spatial and temporal dimensions.

\noindent
\textbf{\textit{Foreground Object Geometry.}}
To understand the dynamics of moving objects, we need to reconstruct them first. 
Given an input video, we first detect, ground, and segment all dynamic objects in every frame using off-the-shelf models~\cite{liu2024grounding,ravi2025sam,cheng2022xmem}. 
For each object instance, we extract an object-centric crop from the first frame (which often has the highest quality), which we feed to InstantMesh~\cite{xu2024instantmesh} for single-image 3D mesh reconstruction. 
Empirically, leveraging pretrained object priors in image-to-3D models yields more reliable meshes than directly leveraging multi-view methods using different frames in our case, which often suffers inconsistent geometry and texture due to flaws in the video generator.

\noindent
\textbf{\textit{Background Scene Geometry.}}
Moreover, to recover a spatio-temporally coherent background scene geometry and camera movements from the template video, we perform 4D reconstruction using ViPE~\cite{huang2025vipe} which masks the foreground dynamic components from the video and leverages bundle adjustment on key frames (See Fig. \ref{fig:perceptionfigs}(a)).
We obtain the 3D background geometry by transforming per-frame metric depth pointmaps to a scene-level world frame and aggregate static background points from all frames. 
From the 4D reconstruction, we also infer the camera poses and rough object positions (see Fig. \ref{fig:perceptionfigs}(b)).
To enhance the quality and reliability of the reconstructed scene geometry, we apply aggressive sub-sampling and filtering to eliminate floating artifacts and invalid points, which often result from {inconsistencies} inherent in the template video.

\noindent
\textbf{\textit{Foreground Object Dynamics.}}
To accurately replicate object dynamics in a physical simulator, it is necessary to determine the object’s initial state, including its position, instantaneous velocity.
We estimate the initial velocity by decomposing it into linear and rotational components.
Specifically, we select two key frames separated by a real-world interval of $\Delta t$. The linear velocity is computed as the estimated 3D displacement vector divided by $\Delta t$ (See Fig. \ref{fig:perceptionfigs}(c)). 
To estimate the rotational velocity, we perform 2D feature matching between object instances in the two frames (See Fig. \ref{fig:perceptionfigs}(d)) using SuperGlue~\cite{sarlin2020superglue}. 
The rotational motion is then isolated by computing a 2D flow field relative to the centroid of the matched feature points (refer to supplemental material for more details).
Finally, we combine the estimated linear and rotational components to derive the per-point initial instantaneous velocity for the object (See Fig. \ref{fig:perceptionfigs}(e)).

\subsubsection{Physical Simulation}
\label{sec:physicalsimulation}

We adopt an MPM-based physical simulator \cite{stomakhin2013material,hu2019taichi} to simulate and generate physically accurate scene dynamics and visual guidance for our subsequent video generation.
A key step is to initialize the scene (geometries and physical properties) in the simulator such that it recreates the intended dynamics in the template video.
Using the outputs from the perception pipeline, the scene initialization includes a few steps, such as determining the simulation domain and estimating physical properties, which we discuss below.

\noindent
\textbf{Simulation Domain.}
To enable stable simulation and seamless mapping from our perception pipeline to the physical simulator, we first need to determine an appropriate simulation domain that is large enough to include the object and possible range of motion of the object and its interactions with the environment, yet the scene domain should also be as small as possible to improve simulation efficiency.
As visualized in Fig.~\ref{fig:perceptionfigs}(f), we first bound the foreground range of dynamics (illustrated as a green box) and the background geometries (illustrated as a red box), and we use an spatial offset coefficient ($C$) to determine a cube (illustrated as a blue box) that appropriately houses the 3D scene centered in the middle. Then, we define the domain box as $[0, 2]$ in $x$, $y$, $z$ directions and scale, rotate and translate all scene geometries and camera parameters with respect to this simulation domain. By doing so, we can therefore determine the appropriate simulation resolution and metric-to-simulation scale ($S$) that will also be used to scale the physics constants such as the gravity value and Young's modulus.

\noindent
\textbf{Physical Property Estimation.}
To enable physically plausible interactions, we initialize the physical properties of objects that influence motion and contact dynamics. Specifically, we employ a large vision-language model (GPT-5 \cite{Gpt-5}) to infer object-specific parameters from the first frame of the template video, guided by a curated text prompt. The model predicts material-related attributes such as density and Young’s modulus.
However, directly estimating numerical values often yields inconsistent or unreliable results. To address this, we design a hierarchical prompting framework that first queries for intermediate material descriptors, including object composition, elasticity or bounce characteristics, and surface roughness; then, these qualitative properties are mapped to corresponding physical parameters to be used in simulation. 
The full prompting pipeline and parameter mapping details are provided in the supplementary material.

\noindent
\textbf{Simulate and Render.}
After initializing the scene with the obtained information, we run the forward MPM physical simulation \cite{stomakhin2013material,hu2019taichi}, obtaining physically plausible high-resolution particle-level trajectories in 3D space.
To obtain explicit guidance signal in pixel space, we render the simulated particle data using Mitsuba~\cite{jakob2022dr} into RGB frames, segmentation masks, and frame-to-frame pixel-to-pixel correspondences using the camera poses estimated from the template video.
These rendered outputs from the physical simulator are then used to guide the video generation.

Note, that we find that the physical simulator cannot directly replace the generator since the simulator's rendered RGB is often unnatural and unrealistic, due to several reasons:
Firstly, the rendered RGB is in a very artificial simulator-like style, and the visual style can be quite unnatural and very different from the style of the intended video background.
Furthermore, physical simulators often cannot handle other factors such as lighting and shadows, and the rendering is also often not in high resolution -- which all affect the quality of the rendered outputs. 
Moreover, there are often inaccuracies and imperfections in the 3D object mesh, which leads to unrealistic rendered video from the physical simulator.
Therefore, these rendered outputs are often not a good replacement for videos generated by video generation models.
Yet, although these renders lack photorealism, they encapsulate faithful motion physics, which is helpful to guide a video generation model,  facilitating the generation of plausible and physically consistent video.
We discuss these further in the next section.

\subsection{Physically-consistent Video Generation}
\label{sec:videogen}

\begin{figure}[t]
    \centering
  \includegraphics[width=1\linewidth]{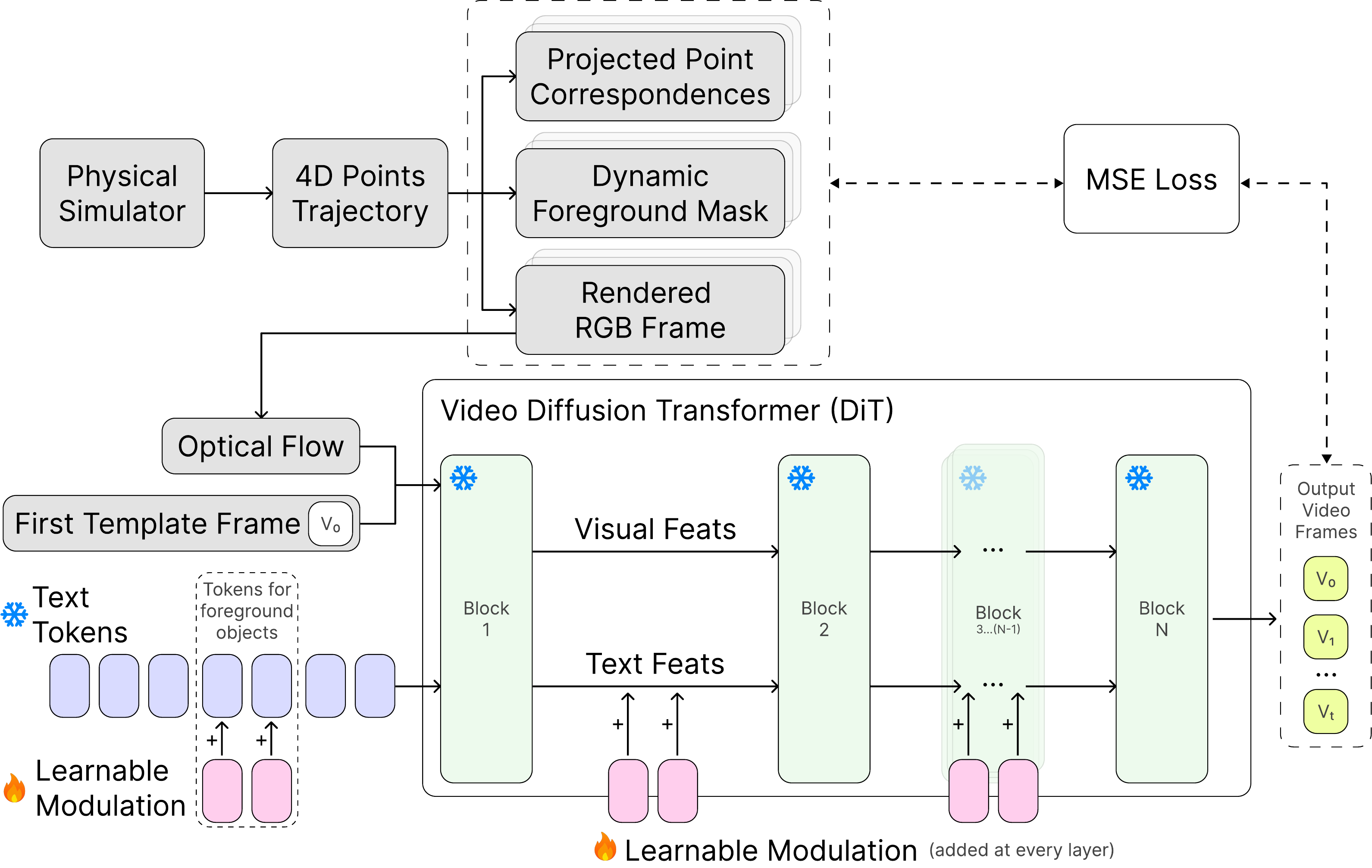}
  \vspace{-6mm}
\caption{Overview of TTCO. To improve the consistencies of textures, during test time, we add learnable zero-initialized embeddings to the text prompt and features, and optimize them with the outputs from the physical simulator. This allows the generated video to adhere to the simulator trajectories and rotations better, thereby improving the texture consistency.
}
\label{fig:testtimeopt}
\vspace{-4.5mm}
\end{figure}

The rendered outputs from the physical simulator, can then be used to guide the video generation.
Such guidance can be performed in several ways, including conditioning the video generation on segmentation masks \cite{akkerman2025interdyn} or depths \cite{alhaija2025cosmos}.
Here, we use an optical flow-conditioned video generation, model Go-with-the-Flow (GwtF) \cite{burgert2025go}, since optical flow  conditioning allows for simultaneously encoding trajectories and rotations, as well as convenient modeling of camera movements with background optical flow.

Specifically, to use GwtF \cite{burgert2025go}, we compute RAFT optical flow \cite{teed2020raft} from two sources: 1) To get physically consistent foreground motion, we compute the optical flow from the simulator-rendered RGB. 
2) To preserve the background scene movements and camera dynamics, we compute optical flow from the template video. 
These flows are fused with the aid of segmentation masks to form a hybrid flow field, preserving real-world motion cues which are difficult to fit and model in the simulator (e.g., water, foliage) and camera movements, while enforcing physics-based constraints for object motion.
Then, the optical flow is used to warp the noise latents (following GwtF \cite{burgert2025go}), which is input into the model along with the prompts and the starting frame of the template video.
Note, that RAFT optical flow is used instead of the simulator's pixel-to-pixel correspondences, since GwtF is trained with RAFT and attains good performance when RAFT optical flow is used.

\subsubsection{Test-time Texture-consistency Optimization}
\label{sec:testtimeopt}

With the above process, we can generate video that follows the general trajectories and movements of the foreground objects.
However, even with accurate motion guidance, existing flow-conditioned video generation models \cite{burgert2025go} still exhibit texture flickering and object appearance drift, often failing to generate objects with consistent textures and colors across the frames. 
For instance, there may be flickering during certain frames, or the textures of certain objects may change unnaturally during rotations.
To address this, we introduce a \textit{test-time texture-consistency optimization} (TTCO) technique to improve the texture-consistency of the generated objects.
See Fig. \ref{fig:testtimeopt} for an overview.

TTCO is a lightweight, test-time procedure that locally adapts the model to maintain texture consistency of foreground objects across frames.
During test-time, we optimize the learnable parameters, so that the generated video follows more accurately the trajectory and rotation of the objects in the physical simulation.
Specifically, this is done by applying a pixel-correspondence loss using the pixel-to-pixel correspondences from the physical simulator, which encourages the pixel-to-pixel movement between frames to follow the physical simulator's foreground.
Note that no  additional data are required for TTCO.

Let $\hat{L}_{\tau}$ denote the latent predicted by the diffusion model at denoising timestep $\tau$. 
Let $\hat{I}_1$ be the first frame of the template video, and let $W_t(\hat{I}_1)$ denote the warping of the first frame to the $t$-th frame using simulator pixel correspondences $  \{(p_{1,j}, q_{t,j})\}_{j \in J} $, where $p_{1,j}$ and $q_{t,j}$ indicate the $j$-th corresponding pixel locations between frames $1$ and $t$. 
The texture consistency loss for the $t$-th frame is defined as:
\vspace{-2mm}

{\small
\setlength{\abovedisplayskip}{0pt}
\setlength{\belowdisplayskip}{0pt}
\begin{equation} 
\mathcal{L}_{\text{tex}}(t)
=
\sum_{j=1}^J
\left\|
    \big[De(h_0(\hat{L}_{\tau}))\big]_{q_{t,j}}
    -
    \big[W_{t}(\hat{I}_1)\big]_{q_{t,j}}
\right\|_2^2,
\label{eq:textureconsistencyloss}
\end{equation}
}
where $h_0(\cdot)$ is the deterministic DDIM-style step mapping to the final denoising iteration \cite{song2021denoising,foo2023distribution}, and $De(\cdot)$ denotes the decoder. The operator $[\cdot]_q$ retrieves the pixel value at location $q$.
We sum up this loss over all frames: $\mathcal{L}_{\text{TTCO}} = \sum_{t=2}^{T} \mathcal{L}_{\text{tex}}(t)$.

In practice, to implement this, we apply the pixel-to-pixel correspondences from the physical simulator to warp the first frame of the template video, creating a texture-consistent target. 
Then, we apply a pixel-wise MSE loss between the generated video and the texture-consistent target, and is applied with the aid of the mask such that only the foreground pixels are considered.
Note, that because the pixel correspondences are often sparse, i.e.~, does not densely cover every pixel of the foreground, we also perform an interpolation operation to compute the dense pixel correspondences of all pixels.
Furthermore, in cases with larger object rotations, the objects quickly rotate till none of the pixels were visible in the original first frame, thus the pixel-to-pixel warping loss cannot be directly enforced; in those cases, we use the pixel values of the rendered reconstructed object in the simulator, which can also facilitate texture consistency in the video.
During this test-time optimization, we focus on sampling the earlier (i.e., noisier) diffusion steps, which we find is important in guiding the generation of textures.

Our TTCO technique aims to improve the texture consistency of the moving foreground objects, thus we want the fine-tuning to be \textit{localized} to target these foreground objects, while preserving the quality of the background. To achieve this, we optimize only foreground-related parameters. We introduce a learnable residual token added to text embeddings for object phrases, as well as feature-wise modulations in DiT layers corresponding to object tokens. By focusing on the text prompt and tokens of the foreground object, we observe that the impact on the video background is greatly minimized as compared to some alternative techniques, e.g., introducing LoRA layers. This provides localized adaptation, improving object texture stability while often leaving the background untouched, which is a crucial requirement for maintaining visual fidelity. 
Our observation that text tokens are strongly related to their corresponding foreground objects also aligns with other recent diffusion-based studies \cite{cai2025ditctrl,garibi2025tokenverse}, further contributing to the growing body of evidence that text-token modulation is a highly effective mechanism for controlling object-specific appearance.
Refer to supplementary material for more details.

\begin{table*}[t]
\centering
\small
\caption{
Quantitative comparison with existing methods for text-to-video generation. 
}
\vspace{-2mm}
\label{tab:quantitative}
\scalebox{0.78}{
\begin{tabular}{@{} l l *{8}{c} @{}}
\toprule
\textbf{Type} & \textbf{Method} &
\multicolumn{2}{c}{\textbf{Motion Controllability}} &
\multicolumn{6}{c}{\textbf{General Video Generation Quality}} \\
\cmidrule(lr){3-4} \cmidrule(lr){5-10}
& &
\makecell{SAM \\ mIoU $\uparrow$} &
\makecell{Corr. Pixel \\ MSE $\downarrow$} &
\makecell{CLIP \\ Text $\uparrow$} &
\makecell{CLIP \\ Img $\uparrow$} &
\makecell{Subject \\ Consistency $\uparrow$} &
\makecell{Background \\ Consistency $\uparrow$} &
\makecell{Motion \\ Smoothness $\uparrow$} &
\makecell{Temporal \\ Flickering $\uparrow$} \\
\midrule
\multirow{5}{*}{\rotatebox[origin=c]{90}{\;Text\mbox{-}based}} 
& \cite{yang2025cogvideox} CogVideoX                 & 0.47 & 0.032 & 0.34 & \textbf{0.99} & 0.93 & 0.95 & 0.98 & 0.97 \\
& \cite{kong2024hunyuanvideo} HunyuanVideo              & 0.46 & 0.017 & \textbf{0.35} & \textbf{0.99} & \textbf{0.95} & \textbf{0.96} & \textbf{0.99} & 0.98 \\
& \cite{li2025pisa} PISA-Base                 & 0.50 & 0.012 & \textbf{0.35} & \textbf{0.99} & \textbf{0.95} & \textbf{0.96} & \textbf{0.99} & \textbf{0.99} \\
& \cite{li2025pisa} PISA-Seg                  & 0.50 & 0.012 & \textbf{0.35} & \textbf{0.99} & \textbf{0.95} & \textbf{0.96} & \textbf{0.99} & \textbf{0.99} \\
& \cite{li2025pisa} PISA-Depth                & 0.51 & 0.017 & \textbf{0.35} & \textbf{0.99} & 0.85 & 0.92 & 0.98 & 0.98 \\
\addlinespace[1pt]
\midrule
\addlinespace[1pt]
\multirow{6}{*}{\rotatebox[origin=c]{90}{Controllable}} 
& \cite{ling2025motionclone} MotionClone               & 0.68 & 0.019 & \textbf{0.35} & 0.97 & 0.87 & 0.92 & 0.97 & 0.94 \\
& \cite{namekata2025sgiv} SG-I2V                    & 0.75 & 0.021 & 0.34 & 0.98 & \textbf{0.95} & 0.95 & 0.97 & 0.94 \\
& \cite{wu2024draganything} DragAnything              & 0.43 & 0.020 & 0.34 & 0.95 & 0.88 & 0.92 & 0.94 & 0.92 \\
& \cite{li2025image} Image Conductor-Object   & 0.61 & 0.022 & 0.34 & 0.95 & 0.84 & 0.92 & 0.93 & 0.90 \\
& \cite{li2025image} Image Conductor-Camera   & 0.55 & 0.023 & \textbf{0.35} & 0.95 & 0.81 & 0.90 & 0.92 & 0.88 \\
\addlinespace[2pt]
& \textbf{Ours (PSIVG)}     & \textbf{0.84} & \textbf{0.007} & \textbf{0.35} & \textbf{0.99} & \textbf{0.95} & \textbf{0.96} & \textbf{0.99} & 0.97 \\
\bottomrule
\end{tabular}
}
\end{table*}

\begin{figure*}[t]
    \centering
  \includegraphics[width=1\linewidth]{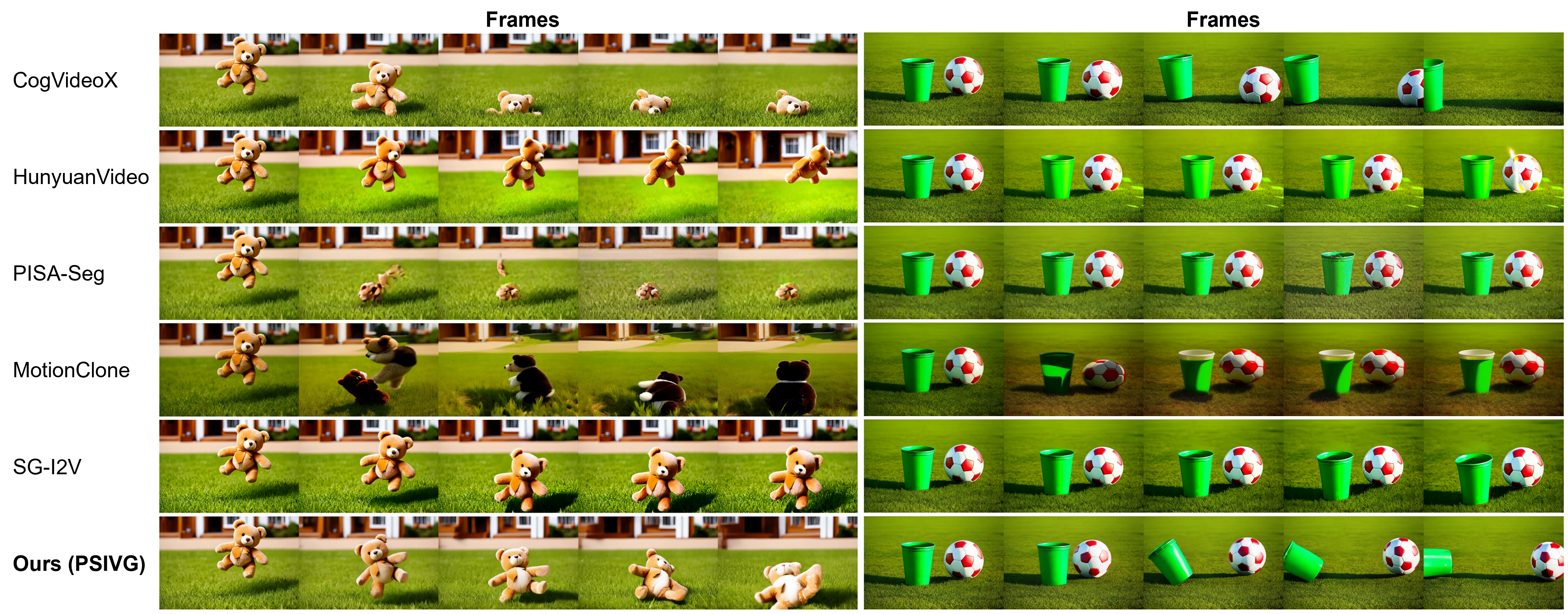}
\caption{
Qualitative comparisons, showing a teddy bear being dropped (left) and objects colliding (right).
See Supp. Mat. for more results.
}
\label{fig:qualitative_results}
\vspace{-4.5mm}
\end{figure*}

\section{Implementation Details}

To generate template videos, we first use SD 3 \cite{esser2024scaling} to generate images with the prompt, and then use CogVideoX-I2V-5B \cite{yang2025cogvideox} or HunyuanVideo-I2V \cite{kong2024hunyuanvideo} with these images and prompts to generate template videos.
During TTCO, we adopt AdamW optimizer, with LR=2e-4. We run it for 50 iterations with our $\mathcal{L}_{TTCO}$ loss.
During TTCO, we also focus on sampling the diffusion steps 700-1000, i.e, the noisier steps, which we find to be important in guiding the generation of textures.
Please refer to the supplementary material for more details.

\section{Experiments}

\subsection{Text-to-Video Generation}

To evaluate our method, we conduct experiments across diverse text prompts automatically generated by an LLM. The prompts include both single- and multi-object scenes, and videos with either static or dynamic camera motion.

\noindent
\textbf{Evaluation Metrics.}
We evaluate our method from two aspects: {motion controllability} and {video quality}.
For motion controllability, we measure how well generated objects follow physically-consistent simulated trajectories using SAM-based mask overlap (\textit{SAM mIoU}) and frame-to-frame pixel correspondence error (\textit{Corr. Pixel MSE}).
For general quality, we assess text alignment via CLIP similarity (\textit{CLIP Text}); and temporal consistency with CLIP image embedding similarity between consecutive frames (\textit{CLIP Img}), and also using VBench \cite{huang2024vbench} metrics (\textit{subject consistency} and \textit{background consistency}, \textit{motion smoothness}, and \textit{temporal flickering}).
See supplementary for more details.

\noindent
\textbf{Baselines.}
First, we compare against open-source T2V models that accept first frame inputs, including CogVideoX \cite{yang2025cogvideox}, HunyuanVideo \cite{kong2024hunyuanvideo}, and variants of PISA \cite{li2025pisa}, which have been trained for better physical accuracy with base finetuning (PISA-Base), segmentation guidance (PISA-Seg), or depth guidance (PISA-Depth).
Next, we also compare with several controllable video generation methods, which are based on masks (DragAnything \cite{wu2024draganything}), object trajectories (Image Conductor-Object \cite{li2025image}) or camera trajectories (Image Conductor-Camera \cite{li2025image}), as well as two training-free motion control methods for controlling movements from reference videos (MotionClone \cite{ling2025motionclone}) or trajectories (SG-I2V \cite{namekata2025sgiv}).
For these controllable baselines that require additional information, we implement them by applying the \textit{conditioning information from our physical simulator}.

\noindent
\textbf{Quantitative Results.}
Please see Tab.~\ref{tab:quantitative} for the results.
We observe that our PSIVG consistently achieves the best performance on the motion controllability metrics (SAM mIoU and Corr. Pixel MSE), indicating a strong ability to maintain physically-consistent and coherent motion trajectories across frames as compared to both text-to-video and controllable video generation baselines. 
Furthermore, although some methods (e.g., PISA-Seg) achieve comparable temporal smoothness and perceptual quality (as seen from motion smoothness and temporal flickering metrics), we find that these models often produce minimal or nearly static movements, where all frames closely resemble the first one, as shown in qualitative results. 
Consequently, despite appearing temporally stable, such methods exhibit poor motion diversity and fail to capture physically realistic dynamics, leading to low scores in the motion accuracy metrics.

\noindent
\textbf{Qualitative Results.}
We provide qualitative comparisons in Fig.~\ref{fig:qualitative_results} across several examples, covering diverse motion scenarios and object interactions.
As shown, our method generates videos that exhibit physically consistent and temporally coherent motion. In contrast, existing text-to-video models (e.g., CogVideoX, HunyuanVideo, PISA-Seg) tend to produce visually appealing but physically implausible motion, such as objects floating in midair, fading away, or jumping around.
Similarly, controllable video generation approaches (e.g.~, MotionClone, SG-I2V) often still struggle with following the trajectory, especially in terms of rotations, and often do not preserve the consistency of the object and background well.
Please see the supplemental material for more visualizations.

\subsection{User Study}

\begin{table}[t]
\centering
\caption{User study results against baseline methods.}
\vspace{-2mm}
\label{tab:userstudy}
\setlength{\tabcolsep}{10pt}
\renewcommand{\arraystretch}{1.0}
\begin{adjustbox}{max width=0.5\linewidth}
\sisetup{
  table-number-alignment = center,
  table-text-alignment   = center,
  table-format           = 2.1,
  detect-weight          = true,
  detect-inline-weight   = math
}
\begin{tabular}{@{} l S @{}}
\toprule
\textbf{Method} & {\textbf{Preference Rate} (\%)} \\
\midrule
CogVideoX            & 7.2 \\
HunyuanVideo         & 4.5 \\
PISA-Seg             & 2.6 \\
SG-I2V               & 2.5 \\
MotionClone          & 0.9 \\
\midrule
\textbf{Ours (PSIVG)} & \textbf{82.3} \\
\bottomrule
\end{tabular}
\end{adjustbox}
\vspace{-1mm}
\end{table}

To further assess physical consistency, we conducted a user study involving 32 participants. 
Each participant was shown sets of videos generated by 5 strong baseline methods and ours, and was asked to select the one that appeared the most physically plausible.
As shown in Tab.~\ref{tab:userstudy}, our method was preferred in 82.3\% of the comparisons, substantially outperforming all baseline models.
This confirms that human evaluators consistently perceive our generated videos as more physically consistent and natural.

\subsection{Ablation Study}

\noindent
\textbf{Impact of TTCO.}
Tab.~\ref{tab:ablation_tto} compares results with and without our proposed TTCO.
Incorporating TTCO notably improves Corr. Pixel MSE, indicating better alignment with pixel-level motion such as rotations, and slightly boosts SAM mIoU, reflecting more accurate object trajectories.
Appearance of the object is also more consistent, as seen from gains in the subject consistency metric.
These results demonstrate the effectiveness of TTCO in enhancing texture consistency.

\begin{figure}[t]
    \centering
  \includegraphics[width=1\linewidth]{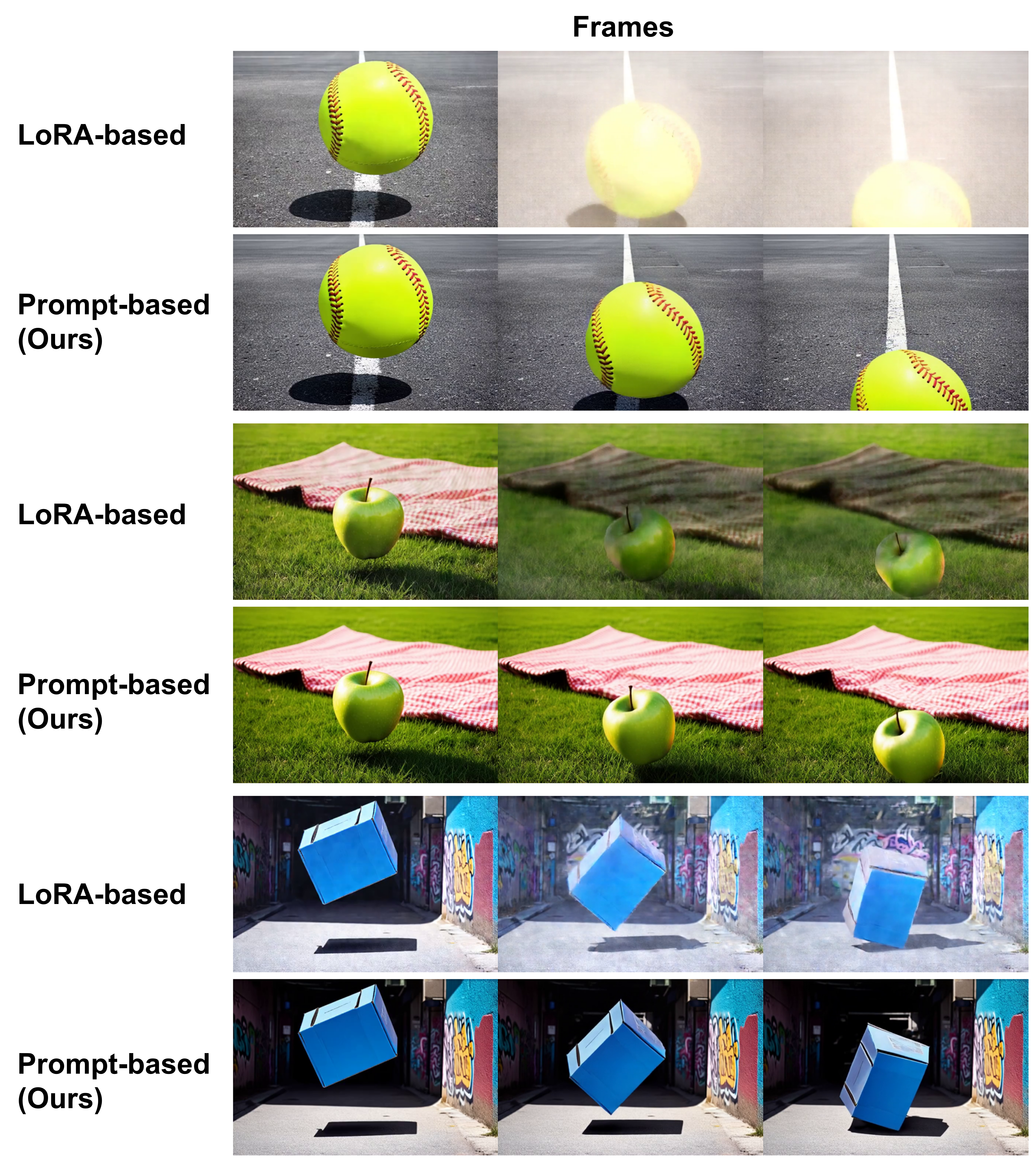}
\caption{
Impact of Prompt-based Optimization Design.
}
\label{fig:ablation_ttco_arch}
\vspace{-3mm}
\end{figure}

\noindent
\textbf{Impact of Prompt-Based Optimization.}
We ablate the efficacy of our prompt-based optimization design for TTCO, comparing it against a baseline that fine-tunes a LoRA at test-time in Fig.~\ref{fig:ablation_ttco_arch}.
We observe that the LoRA-based design often degrades video quality, particularly in the background, whereas  our prompt-based method consistently yields higher quality results. 
We attribute this improvement to the lightweight, localized nature of prompt-based optimization, which preserves global visual consistency while refining object-level details.
Besides, we also tested a design where we directly optimized the object-specific spatio-temporal tokens instead of text tokens, but found that it often produced grid-like artifacts. In comparison, modulating text prompts and tokens is lightweight and works well.

\begin{table}[t]
\centering
\caption{Impact of Test-Time Optimization (TTCO).}
\vspace{-2mm}
\label{tab:ablation_tto}
\setlength{\tabcolsep}{5pt}
\renewcommand{\arraystretch}{1.0}
\begin{adjustbox}{max width=0.85\linewidth}
\sisetup{
  table-number-alignment = center,
  table-text-alignment   = center,
  detect-weight          = true,
  detect-inline-weight   = math
}
\begin{tabular}{@{} l
                S[table-format=1.2]
                S[table-format=1.3] 
                S[table-format=1.2] @{}}
\toprule
\textbf{Setting} & {\textbf{SAM mIoU}~$\uparrow$} & {\textbf{Corr.\ Pixel MSE}~$\downarrow$} & {\textbf{Subj. Consis.}~$\uparrow$} \\
\midrule
w/o TTCO        & 0.82 & 0.009  & 0.93 \\
w/ TTCO (ours)  & \textbf{0.84} & \textbf{0.007}  &  \textbf{0.95}  \\
\bottomrule
\end{tabular}
\end{adjustbox}
\vspace{-3mm}
\end{table}

\section{Conclusion}
We present \textbf{PSIVG}, a physical simulator in-the-loop video generation framework that effectively integrates physical simulation into diffusion-based video generation, and enhances texture consistency via TTCO.
Extensive experiments demonstrate that PSIVG produces videos with superior physical realism and visual quality compared to existing methods.

\noindent
\textbf{Limitations.} 
While effective, our method faces several limitations: 
(1) Reliance on MPM \cite{stomakhin2013material} limits our ability to handle complex agents, such as humans or vehicles, and articulated structures. 
(2) Limitations in perception quality during initial object reconstruction. 
(3) We inherit the generative limitations of the GwtF video model \cite{burgert2025go}, e.g., difficulties in generating very small or thin objects. 
Refer to supplementary material for more details.

\noindent
\textbf{Acknowledgments.} This work was supported by the Saarbrücken Research Center for Visual Computing, Interaction and Artificial Intelligence (VIA).



\end{document}